\documentclass{article}
\usepackage{ijcai25}

\usepackage{times}
\usepackage{soul}
\usepackage{url}
\usepackage[hidelinks]{hyperref}
\usepackage[utf8]{inputenc}
\usepackage{graphicx}
\usepackage{amsthm}
\usepackage{multirow}
\usepackage{amsmath,amsfonts,amssymb}
\usepackage{amsthm}
\usepackage{booktabs}
\usepackage{algorithm}
\usepackage{algorithmic}
\usepackage[switch]{lineno}
\usepackage{comment}
\usepackage{pifont}
\usepackage{multirow}
\usepackage{xcolor}
\usepackage{caption} 
\title{Filling the Missings: Spatiotemporal Data Imputation by Conditional Diffusion}

\author{Wenying He$^1$ 
\and
Jieling Huang$^1$
\and
Junhua Gu$^{1,*}$
\and
Ji Zhang$^3$
\textnormal{and}
Yude Bai$^{2,*}$\\
\affiliations
$^1$School of Artificial Intelligence, Hebei University of Technology, Tianjin, China \\
$^2$School of Software, Tiangong University, Tianjin, China \\
$^3$University of Southern Queensland, Queensland, Australia \\
\emails
\ {\{hwying1234, jhgu\}}@hebut.edu.cn,
202332805027@stu.hebut.edu.cn,
Ji.Zhang@unisq.edu.au,
baiyude@tiangong.edu.cn
}

\begin{document}

\maketitle

\renewcommand{\thefootnote}{\fnsymbol{footnote}}
\footnotetext[1]{Corresponding author: Junhua Gu, Yude Bai.}
\renewcommand{\thefootnote}{\arabic{footnote}}

\begin{abstract}
   Missing data in spatiotemporal systems presents a significant challenge for modern applications, ranging from environmental monitoring to urban traffic management. The integrity of spatiotemporal data often deteriorates due to hardware malfunctions and software failures in real-world deployments. Current approaches based on machine learning and deep learning struggle to model the intricate interdependencies between spatial and temporal dimensions effectively and, more importantly, suffer from cumulative errors during the data imputation process, which propagate and amplify through iterations. To address these limitations, we propose CoFILL, a novel Conditional Diffusion Model for spatiotemporal data imputation. CoFILL builds on the inherent advantages of diffusion models to generate high-quality imputations without relying on potentially error-prone prior estimates. It incorporates an innovative dual-stream architecture that processes temporal and frequency domain features in parallel. By fusing these complementary features, CoFILL captures both rapid fluctuations and underlying patterns in the data, which enables more robust imputation. The extensive experiments reveal that CoFILL's noise prediction network successfully transforms random noise into meaningful values that align with the true data distribution. The results also show that CoFILL outperforms state-of-the-art methods in imputation accuracy. The source code is publicly available at https://github.com/joyHJL/CoFILL.
\end{abstract}


\section{Introduction}\label{sec:intro}

Spatiotemporal data integrates temporal dynamics with spatial relationships to represent complex real-world phenomena. This data type has proven valuable across diverse domains, from environmental science to urban systems~\cite{ref1,ref2,ref4}. Despite its importance, the collection of spatiotemporal data faces significant challenges in real-world deployments. Sensor malfunction, system instability, and environmental interference often lead to missing or corrupted measurements. These data quality issues compromise the reliability of information extracted from spatiotemporal systems and impair the performance of downstream applications~\cite{ref6,ref7,ref45}.

Various methods have been proposed to handle missing data in spatiotemporal datasets. Traditional approaches rely on statistical and data analytic techniques, and make assumptions about the underlying probability distributions and use techniques such as linear interpolation~\cite{ref8}, expectation-maximization (EM) algorithm~\cite{ref9,ref10}, or k-nearest neighbors (KNN)~\cite{ref12} to estimate missing data. Their effectiveness is constrained by their inherent mathematical foundations, which rely on simplified models of relationships between variables. When applied to real-world data, where relationships between spatial and temporal components often follow complex, nonlinear patterns, these methods fail to reconstruct the true data distribution accurately.

Deep learning methods have also been applied for spatiotemporal data imputation to better process the inherent complexity in spatiotemporal relationships. Recurrent neural networks (RNN)~\cite{ref14} model temporal relationships through hidden state propagation, while graph neural networks (GNN)~\cite{ref15,ref3} learn spatial dependencies through node-level representations. These methods, however, face fundamental limitations. RNN~\cite{ref32} operate through an autoregressive mechanism, which causes early prediction errors to propagate through subsequent time steps. GNN~\cite{ref48} depend on features from neighboring nodes, allowing errors to disseminate across the graph structure. The recursive nature of their processing leads to error accumulation over time. Their reliance on historical data also constrains their ability to capture the full range of possible values during imputation.

Alternatively, data imputation can be treated as a generative task~\cite{ref16}. Recent work has shown that diffusion probabilistic models (DDPM)~\cite{ref19} and generative adversarial networks (GAN)~\cite{ref20}. However, current DDPM-based methods face two key challenges. First, they fail to model both temporal and spatial aspects simultaneously~\cite{ref4} - for example, Tashiro et al.~\cite{ref21} consider only temporal features, ignoring crucial spatial relationships. Second, they rely on problematic pre-imputation approaches - Liu et al.~\cite{ref22} use linear interpolation that creates artificial fluctuations, causing generated data to deviate from actual patterns. Our work addresses these limitations by developing a model that integrates both spatial and temporal information while avoiding error-prone pre-imputation steps.

In this paper, we introduce CoFILL (\textbf{\underline{Co}}nditional Di\textbf{\underline{f}}fus\textbf{\underline{i}}on Model based on Tempora\textbf{\underline{l}}-Frequency Spatiotempora\textbf{\underline{l}} Imputation), a novel approach that addresses the limitations of current methods. Our model enhances imputation by utilizing temporal measurements from sensors and geographical connections between monitoring locations. The temporal data captures changing patterns over time, while spatial relationships reveal location interdependencies. The architecture features modules for temporal and frequency domain feature extraction to capture rapid changes and persistent patterns. A cross-attention mechanism combines these features into rich conditional information. This information guides the noise prediction network to transform random noise into meaningful data values through iterative denoising, which enables robust imputation by learning the underlying data distribution rather than direct estimation.

Our contributions are summarized as follows:
\begin{itemize}
\item We propose a spatiotemporal imputation framework, CoFILL, based on diffusion models. It applies the non-recursive property of diffusion models to effectively reduce error accumulation during the imputation process.
\item We develop a dual-stream feature processing architecture to model complete spatiotemporal dynamics. We extract temporal and frequency information through separate pathways, then integrate them by cross-attention to capture both rapid fluctuations and underlying patterns.
\item We conduct extensive experiments on three real-world datasets to validate our approach. The results demonstrate that CoFILL outperforms current state-of-the-art methods in spatiotemporal data imputation tasks.
\end{itemize}


\section{Related Works}\label{sec:related}

\subsection{Spatiotemporal Data Imputation}\label{sec:related1}
Methods for spatiotemporal data imputation mainly fall into three main categories: statistical approaches, machine learning techniques, and deep learning models~\cite{ref47}. Statistical methods, such as mean and median imputation~\cite{ref23}, assume the data follows specific probability distributions, which is too restrictive for real-world data. Machine learning approaches provide more flexibility like random forest (RF)~\cite{ref24} and matrix factorization~\cite{ref25}. Other machine learning methods such as Bayesian Gaussian process tensor completion (BGCP)~\cite{ref26} and time-regularized tensor factorization (TRTF)~\cite{ref27} show promise in processing complex data. But they still struggle with nonlinear relationships and scalability to large datasets.
Deep learning methods, such as recurrent neural networks (RNN)~\cite{ref29} and their variants like GRU-D~\cite{ref30}, excel at learning temporal patterns but accumulate errors over prediction sequences. Convolutional neural networks, such as TimesNet~\cite{ref31}, only analyze time series in Euclidean space using Fourier transforms. Graph neural networks~\cite{ref33,ref5} improve imputation accuracy by modeling spatial relationships, but their reliance on fixed graph structures reduces their adaptability to new scenarios. These existing approaches cannot fully capture the nonlinear dependencies and complex patterns inherent in large-scale spatiotemporal data.

\subsection{Diffusion Models for Imputation}\label{sec:related2}
Compared with traditional methods and deep learning methods, the generative data imputation method by diffusion models generates high-quality data. The conditional score-based diffusion model (CSDI)~\cite{ref21} uses a subset of observed data as conditions to generate residual values. Meanwhile, PriSTI~\cite{ref22} introduces spatiotemporal relationships and attention mechanisms to achieve better imputation results. GP-VAE~\cite{ref34} models temporal patterns using Gaussian processes in latent space, while E2GAN~\cite{ref35} offers a simplified approach to generate plausible values for time series data. However, current diffusion models struggle to effectively extract and use conditional information - the known patterns and relationships in the existing data that guide the generation of missing data. This limitation impacts imputation accuracy because of the generative values that fail to maintain the underlying temporal and spatial dependencies of the data.

\subsection{Temporal and Frequency Domain Feature}\label{sec:related3}
Feature extraction plays a vital role in spatiotemporal data imputation. Temporal feature extraction through recurrent structures like GRU-D~\cite{ref30} captures immediate state changes but cannot handle variations at different time scales. Meanwhile, frequency domain approaches via Fourier transforms, such as TimesNet~\cite{ref31}, identify periodic and cyclical patterns but miss important transient changes. GNN-based methods~\cite{ref33} focus on modeling spatial relationships but overlook these temporal dynamics.
We hope to address these issues by integrating temporal and frequency domain features through a cross-attention mechanism. The temporal features capture immediate state transitions, while frequency features reveal long-term trends.


\section{Problem Formulation}\label{sec:preliminary}
Consider a spatiotemporal system with $N$ observation nodes (such as air monitoring stations or traffic sensors) measured over $L$ time steps. The measurements form a sequence $X_{1:L} = {X_1, X_2, \dots, X_L} \in \mathbb{R}^{N \times L}$, where $X_l \in \mathbb{R}^N$ represents the values observed at time step $l$. The spatial relationships between nodes are represented by an adjacency matrix $A \in \mathbb{R}^{N \times N}$, where $A_{i,j} = 1$ indicates a connection between nodes $i$ and $j$, and $A_{i,j} = 0$ indicates no connection.
In real-world deployments, some observations may be missing due to sensor failures or system errors. We use a binary mask $M \in {0,1}^{N \times L}$ to indicate the presence or absence of measurements, where $m_{i,l} = 1$ denotes a valid observation at node $i$ and time $l$, while $m_{i,l} = 0$ indicates missing data.
The spatiotemporal imputation problem aims to reconstruct missing data in $X$ given the observed measurements and the adjacency matrix $A$. The goal is to develop an imputation method that preserves both the temporal evolution of measurements and their spatial relationships across nodes.


\section{Methodology}\label{sec:method}

\begin{figure}[t]
	\centerline{\includegraphics[width=\linewidth]{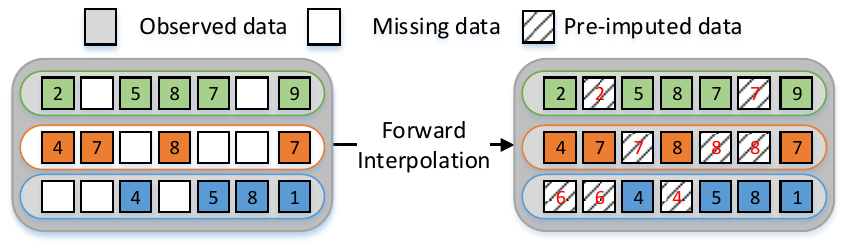}}
	\caption{Pre-imputing by forward interpolation.
    }    
	\label{fig:fig1preimpute}
\end{figure}

\begin{figure*}[t]
	\centerline{\includegraphics[width=0.95\linewidth]{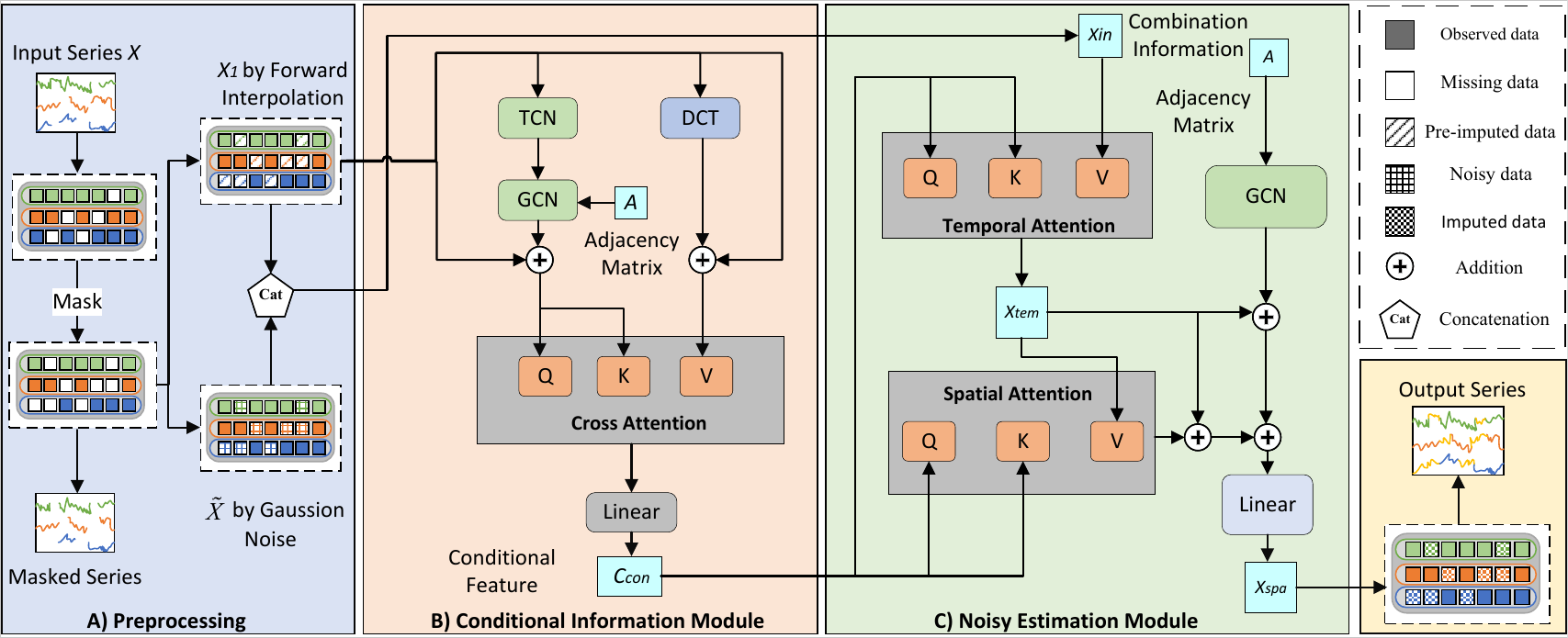}}
	\caption{The framework of CoFILL. 
    }
	\label{fig:fig2overview}
\end{figure*}

\subsection{Preprocessing of CoFILL}\label{sec:method1}

CoFILL performs a two-step preprocessing on the input series to ensure high-quality inputs for training. We first mask some of the observed values in the original data based on a random masking strategy. This not only simulates missing patterns but also enhances the model’s generalization to diverse scenarios~\cite{ref21}. Two sets of initial imputed data are then generated from the masked data. The first set uses forward interpolation, as shown in Figure~\ref{fig:fig1preimpute}. It fills in missing data by using the known values from the previous time step, thus preserving the continuity and periodicity of the spatiotemporal data. This also reduces information loss due to over-smoothing. The second set fills the missing data with Gaussian noise. We introduce noise consistent with the original data distribution to enhance data diversity and introduce appropriate randomness for the model~\cite{ref34,ref49}. These two imputed data sets are adopted as inputs for subsequent modules, which further supports CoFILL in learning spatiotemporal dependencies and missing data distributions.

\subsection{Conditional Information Module}\label{sec:method2}

We extract features from the imputed data sets of preprocessing to provide effective conditional information for the noise prediction network. These features, by identifying the spatiotemporal dependencies from both the observed and pre-imputed data, guide the network to predict noise more accurately. The process works as follows. Initially, we map the imputed data $ X_1 $ to a latent space $ H_{in} = \text{Conv}(X_1) \in \mathbb{R}^{d \times N \times L} $, where $ \text{Conv}(\cdot) $ refers to a $ 1 \times 1 $ convolution operation, and $ d $ denotes the channel size. Next, we extract features from both the temporal and frequency domains, and the combined feature is obtained through cross-attention.

In the temporal domain, data reflects instantaneous changes, which is crucial for short-term prediction. The temporal feature extraction module separately obtains both temporal and spatial dependencies. We map $ X_1 $ to the latent space $ H_{in} $ and pass it through a Temporal Convolutional Network (TCN) to learn temporal dependencies. The TCN employs padded 1-dimensional gated causal convolutions, ensuring that the output and input maintain the same shape. The convolution kernel $ \Gamma_{\tau} \in \mathbb{R}^{K \times C_{in} \times C_{out}} $ generates outputs $ P, Q \in \mathbb{R}^{C_{out} \times N \times L} $ of the same shape. The definition and specific formulas of the TCN structure can be, respectively, expressed as follows:

\begin{equation}
\bar{H}_{in} = \Gamma_{\tau} (H_{in}) = P \odot \sigma(Q) \in \mathbb{R}^{C_{out} \times N \times L}
\end{equation}
\begin{equation}
\begin{aligned}
\bar{H}_{in} =\ & Dropout\left(Chomp\left(Conv2D_{(k\times k)}^{(dilation)}(H_{in})\right)\right) \\
& + H_{in}
\end{aligned}
\end{equation}

\noindent where $ \odot $ represents the Hadamard product, and $ \sigma $ is the sigmoid activation function, introducing nonlinearity and acting as a gating mechanism. We denote the output of the TCN as $\bar{H}_{in}\in R^{d\times N\times L}$, the convolution kernel size is $k$, the dilation size is 1, which indicates short-term feature extraction, and the convolution is expanded along the time axis. We use a Graph Convolutional Network (GCN) to learn spatial dependencies. The definition and specific formulas of the GCN structure can be, respectively, expressed as follows:

\begin{equation}
\tilde{H}_{in} = \Gamma_G (\bar{H}_{in}) = \sigma(\varphi(A_{gcn}, \bar{H}_{in}) W_i)
\end{equation}
\begin{equation}
\begin{aligned}
\tilde{H}_{in} = \text{ReLU}\Big(&\text{Conv2D}\Big( \text{Concat}\Big( \bar{H}_{in}, \\
& \big\{ (\bar{H}_{in} A_{gcn})^{(o)} \big\} \Big) \Big) \Big)
\end{aligned}
\end{equation}
\noindent where $ W_i \in \mathbb{R}^{C_{in}^g \times C_{out}^g} $ is a learnable parameter, $ \sigma $ is the sigmoid activation function, and $ \varphi(\cdot) $ represents the aggregation function that determines how neighboring node features are aggregated into the target node. The graph-normalized adjacency matrix is $ A_{gcn} = D^{-1/2} (A + I) D^{-1/2} \in \mathbb{R}^{N \times N} $, where $ I $ is the identity matrix, and $ D $ is the diagonal matrix with $ D_{ii} = \sum_j \partial(A + I)_{ij} $. $A_{gcn}$ is the static adjacency matrix, $\mathrm{o}$ is the convolution order of the graph, and $\mathrm{Conv2D}$ indicates that convolution is used for channel fusion. We denote the output of both TCN and GCN as $ \tilde{H}_{in}\in R^{d\times N\times L} $. By processing dynamic characteristics in the temporal dimension with the TCN, we extract the temporal changes in the data. At the same time, we adopt the GCN to learn the relationships between nodes in the spatial dimension. This combination effectively identifies the complex changes in spatiotemporal data over time.

In the frequency domain, data reveals periodicity and trends, which are essential for long-term forecasting and recognizing periodic patterns. Many time series models fail to account for the frequency impact, which leads to an inability to extract the intrinsic features of the series. To this end, we apply the Discrete Cosine Transform (DCT) to extract frequency domain features $ \hat{H}_{in} $. Unlike Fourier transforms, DCT avoids the Gibbs phenomenon. It also provides superior energy compaction compared to Wavelet transforms~\cite{ref36}. Its specific formula is as follows:
\begin{equation}
\hat{H}_{in}=\sum_{t=0}^{T-1}H_{in}\cdot\cos\left[\frac{\pi}{T}\left(t+\frac{1}{2}\right)m\right],m=0,1,\cdots,T-1
\end{equation}

\noindent where $H_{in}\in R^{d\times N\times L}, \hat{H}_{in}\in R^{d\times N\times L}$, $\mathrm{T}$ represents the time step. As m→0, the corresponding DCT component captures low-frequency signals, reflecting long-term stable trends. Conversely, as m→T, it captures high-frequency components, indicating periodic structures. Together, the different frequency components in DCT represent both the long-term trends and periodic patterns within the sequence.

As shown in Figure~\ref{fig:fig2overview} B), we apply cross-attention to combine the temporal-domain feature $ \tilde{H}_{in} $ and the frequency-domain feature $ \hat{H}_{in} $ into a mixed feature representation. Temporal and frequency-domain features offer different perspectives. Temporal features reveal short-term behavior, while frequency-domain features identify long-term trends. The fusion of both perspectives enables the system to process multi-scale data within a unified framework. The temporal feature $ Q $ retains time sequence, local changes, and spatial features. It encodes immediate information along the time axis. The frequency-domain feature $ KV $ focuses on global periodicity and trends, which reduces noise interference while preserving key signal characteristics. The combination of both captures both local temporal information and global periodic trends. The time-frequency domain attention weights and the final conditioning information $ C_{con} $ are defined as:

\begin{align}
Q_{STF} &= \tilde{H}_{in} \cdot W_{STF}^Q, \nonumber \\
K_{STF} &= \hat{H}_{in} \cdot W_{STF}^K, \nonumber \\
V_{STF} &= \hat{H}_{in} \cdot W_{STF}^V
\end{align}

\begin{align}
C_{con} &= \text{Attn}_{stf} (Q_{STF}, K_{STF}, V_{STF}) \nonumber \\
&= \text{softmax}\left(\frac{Q_{STF} K_{STF}^T}{\sqrt{d}}\right) \cdot V_{STF}
\end{align}

\subsection{Noisy Estimation Module}\label{sec:method3}

The noise prediction network $\varepsilon$ in this module is employed during the reverse denoising process to progressively sample noise, as illustrated in Figure~\ref{fig:fig2overview} C). It uses conditional information $C_{con}$ to guide the extraction of spatiotemporal dependencies in the noise data $X_{in}$. Additionally, it leverages an attention mechanism to mitigate the randomness introduced by Gaussian noise injected into the data distribution. The model accepts three inputs, including noise information $X_{in} = (X_1 \parallel \tilde{X}^t)$, conditional information $C_{con}$, and the adjacency matrix $A$. It first generates temporal features $X_{tem}$ through a temporal attention module $\Upsilon_T (\cdot)$. The data is then passed into a spatial attention module $\Upsilon_S (\cdot)$, equipped with a GNN, to produce spatial features $X_{spa}$. This process is formalized as follows:

\begin{align}
X_{tem} = \Upsilon_T (X_{in}) = \text{Attn}_{tem} (X_{in})
\end{align}

\begin{align}
X_{spa} &= \Upsilon_S (X_{tem}, A) \nonumber \\ 
&= \text{MLP}(\text{Norm}(\text{Attn}_{spa} (X_{tem})) + X_{tem}) \nonumber \\ 
&\quad + \text{Norm}(G(X_{tem}, A) + X_{tem})
\end{align}

$\text{Attn}_{tem} (\cdot)$ and $\text{Attn}_{spa} (\cdot)$ represent temporal and spatial attention, respectively. The mechanisms are the dot-product multi-head self-attention from the transformer architecture. We also put conditional information into the attention calculations to reduce the noise accumulated during the diffusion steps. Specifically, for $\text{Attn}_{tem} (\cdot)$, the dot-product attention $\text{Attn}_{tem} (Q_T, K_T, V_T) = \text{softmax}((Q_T K_T^T)/\sqrt{d}) \cdot V_T$ is rewritten as $\text{Attn}_{tem} (A_T, V_T) = A_T \cdot V_T$, where $A_T = \text{softmax}((Q_T K_T^T)/\sqrt{d})$ denotes the attention weights. The attention weight $A_T$ is calculated by the conditional information $C_{con}$. We set the inputs $Q_T$, $K_T$, and $V_T$ as follows:

\begin{align}
Q_T = C_{con} \cdot W_T^Q, \quad K_T = C_{con} \cdot W_T^K, \quad V_T = X_{in} \cdot W_T^V
\end{align}

\noindent where $W_T^Q, W_T^K, W_T^V \in \mathbb{R}^{d \times d}$ are learnable projection parameters. Similarly, the spatial attention $\text{Attn}_{spa} (A_S, V_S)$ computes the attention weight in the same manner:

\begin{align}
Q_S = C_{con} \cdot W_S^Q, \quad K_S = C_{con} \cdot W_S^K, \quad V_S = X_{tem} \cdot W_S^V
\end{align}

The noise prediction network consists of multiple stacked attention layers that progressively refine spatial representations. Each layer processes features through a gated activation unit, producing spatial features $X_{spa}$ that split into two paths. The primary path forms residual connections feeding into subsequent layers, while a secondary path creates skip connections that aggregate features across network depths. The network combines these accumulated features and processes them through two 1-dimensional convolution layers, which contains only the values of the interpolation targets.

\subsection{Training and Prediction}\label{sec:method4}

We apply a random masking strategy, during the training, to the input observations $X$ to derive imputation targets $\tilde{X}$, while the remaining observations are used to extract pre-imputation features $X_1$ as conditional information. Similar to CSDI, we implement masking strategies including point strategy, block strategy, and hybrid strategy. As shown in Algorithm~\ref{alg:alg1traning}, the noise prediction model of CoFILL is trained by iterating over the observed data. Specifically, once obtaining the training imputation targets $\tilde{X}$ and conditional information $X_1$, the training objective for spatiotemporal imputation is:

\begin{align}
L(\theta) = \mathbb{E}_{\tilde{X}^0 \sim q(\tilde{X}^0), \epsilon \sim \mathcal{N}(0, I)} \left\| \epsilon - \epsilon_{\theta} (\tilde{X}^t, X_1, A, t) \right\|^2
\end{align}

\begin{algorithm}[t]
\caption{Training of CoFILL}\label{alg:alg1traning}
\renewcommand{\algorithmicrequire}{\textbf{Input:}}
\renewcommand{\algorithmicensure}{\textbf{Output:}}
\begin{algorithmic}
\REQUIRE Incomplete observed data $X$, the adjacency matrix $A$, the number of iterations $N$, the number of diffusion steps $T$, noise levels sequence $\bar{\alpha}_t$.
\ENSURE Optimized noise prediction model $\epsilon_\theta$
\FOR{$i=1$ to $N$}
    \STATE $X_{\text{mask}} \leftarrow \text{Mask}(X)$;
    \STATE $X_1 \leftarrow \text{Interpolate}(X_{\text{mask}})$;
    \STATE Sample $t \sim \text{Uniform}(\{1, \cdots, T\})$, $\epsilon \sim \mathcal{N}(0, I)$;
    \STATE $\tilde{X}^{t} = \sqrt{\bar{\alpha}_t} X^{0} + \sqrt{1 - \bar{\alpha}_t} \epsilon$;
    \STATE Updating the gradient $\nabla_\theta  \left\| \epsilon - \epsilon_\theta (\tilde{X}^{t}, X_1, A, t) \right\|^2$
\ENDFOR
\end{algorithmic}
\end{algorithm}

In the imputation, we deploy the trained noise prediction model $\epsilon_\theta$ and the available observation masks $M$ to perform imputation. The imputation targets consist of all missing data in the spatiotemporal data. As shown in Algorithm~\ref{alg:alg2imputation}, the trained model will fill in the missing data in the spatiotemporal data by performing reverse diffusion.

\begin{algorithm}[t]
\caption{Imputation of CoFILL}\label{alg:alg2imputation}
\renewcommand{\algorithmicrequire}{\textbf{Input:}}
\renewcommand{\algorithmicensure}{\textbf{Output:}}
\begin{algorithmic}
\REQUIRE A sample of incomplete observed data $X$, the adjacency matrix $A$, the number of diffusion steps $T$, the optimized noise prediction model $\epsilon_\theta$.
\ENSURE Missing data of the imputation target $X_{\text{out}}$
    \STATE $X_1 \leftarrow \text{Interpolate}(X)$
    \STATE Set $\tilde{X}^{t} \sim \mathcal{N}(0, I)$
    \FOR{$t = T$ to $1$}
        \STATE $\mu_\theta (\tilde{X}^{t}, X_1, A, t) \leftarrow \frac{1}{\sqrt{\bar{\alpha}_t}} (\tilde{X}^{t} - \frac{\beta_t}{\sqrt{1 - \bar{\alpha}_t}} \epsilon_\theta (\tilde{X}^{t}, X_1, A, t))$
        \STATE $\tilde{X}^{t-1} \leftarrow \mathcal{N}(\mu_\theta (\tilde{X}^{t}, X_1, A, t), \sigma_t^2 I)$
    \ENDFOR
\end{algorithmic}
\end{algorithm}


\section{Experiments}\label{sec:experiment}

\subsection{Datasets}\label{sec:experiment1data}

We use three spatiotemporal datasets for imputation algorithms. The AQI-36 dataset~\cite{ref37} serves as a good benchmark due to its inherent missing data patterns and complex spatiotemporal dependencies. The hourly PM2.5 measurements across Beijing's 36 monitoring stations over a full year handle both short-term fluctuations and long-term season trends. The traffic datasets, METR-LA and PEMS-BAY~\cite{ref2}, present different scales to test our model's capabilities. METR-LA's 207 sensors and PEMS-BAY's 325 sensors represent large-scale networks with dense spatial correlations. Their five-minute sampling frequency verifies model ability to face rapid temporal variations, while their different geographic contexts - Los Angeles and San Francisco Bay Area - validate results across urban traffic.


\subsection{Baselines for Comparison}\label{sec:experiment2baseline}

\textbf{Traditional Statistical Baselines.}
The fundamental statistical approaches include Mean, which imputes values using historical node averages, and Daily Average (DA), which uses time-specific averages. We also include K-Nearest Neighbors (KNN)~\cite{ref11,ref12}, which leverages geographical proximity for imputation, and Linear interpolation for time series reconstruction. For robust sequential data processing, we incorporate Kalman Filter (KF), which estimates missing data through state-space modeling.

\textbf{Matrix and Tensor-Based Baselines.}
These intermediate approaches bridge classical statistics and modern machine learning by decomposing complex data structures. MICE~\cite{ref38} uses conditional specifications with chained equations for multivariate imputation. Vector Autoregression (VAR) predicts single-step values through temporal dependencies. Time-Regularized Matrix Factorization (TRMF)~\cite{ref39} specializes in temporal data patterns, and Bayesian Augmented Tensor Factorization (BATF)~\cite{ref40} integrates spatiotemporal knowledge through tensor decomposition.

\textbf{Deep Learning Baselines.}
For probabilistic modeling with uncertainty quantification, we employ V-RIN~\cite{ref43} and GP-VAE~\cite{ref34}. To leverage adversarial learning, we use rGAIN~\cite{ref42}, which enhances GAIN through bidirectional processing. We further employ BRITS~\cite{ref41} for RNN-based temporal modeling and GRIN for graph neural network processing with bidirectional GRU. Our evaluation also encompasses recent diffusion-based methods: CSDI~\cite{ref21}, which captures feature interactions through transformers, and PriSTI~\cite{ref22}, which implements spatiotemporal attention mechanisms.


\subsection{Evaluation Metrics}\label{sec:experiment3metric}

We investigate the performance by three complementary metrics - Mean Absolute Error (MAE), Mean Squared Error (MSE), and Continuous Ranked Probability Score (CRPS)~\cite{ref44}. MAE and MSE provide measures of imputation accuracy by calculating the distance between predicted values and ground truth. CRPS evaluates the quality of probabilistic predictions by measuring how well the predicted probability distributions align with given values. 


\subsection{Experimental Design}\label{sec:experiment4design}
To ensure reproducibility and meaningful evaluation, we establish a methodical data partitioning strategy that respects the temporal characteristics of our datasets. The AQI-36 dataset requires careful temporal consideration due to its seasonal nature. We distribute the data across seasons by selecting March, June, September, and December for testing, which captures seasonal variations throughout the year. The validation set draws from the final 10\% of data in February, May, August, and November to maintain seasonal representation. The remaining months form the training set. 

To evaluate model performance under different missing data scenarios, we design two types of data corruption schemes. For AQI-36, we implement a simulated failure (SF) pattern that replicates real-world sensor malfunction distributions. For PEMS-BAY and METR-LA traffic datasets, we create controlled missing data scenarios through mask matrices. These scenarios include random point missing (Point), where we mask 25\% of observations uniformly at random, and structured block missing (Block), which combines 5\% random masking with continuous missing segments. These segments span 1 to 4 hours per sensor and occur with 0.15\% probability, simulating extended sensor outages.

Our implementation runs on an NVIDIA RTX 4090 GPU with 24GB VRAM. We optimize the model using Adam with cosine annealing learning rate decay. The learning rate starts at $10^{-3}$ and decays to $10^{-5}$. Table~\ref{tab:tab1parameter} details the complete hyperparameter configuration for each dataset. To establish statistical significance, we conduct five independent runs with different random seeds for each experiment.

\begin{table}[h!]
\footnotesize
\setlength{\tabcolsep}{3pt} 
\centering
\begin{tabular}{c|c|c|c}
\hline
 & \textbf{AQI-36} & \textbf{METR-LA} & \textbf{PEMS-BAY} \\
\hline
Batch size & 16 & 16 & 16 \\
Time length \textit{L} & 36 & 24 & 24 \\
Epochs & 200 & 300 & 300 \\
Learning rate & 0.001 & 0.001 & 0.001 \\
Channel size \textit{d} & 64 & 64 & 64 \\
Noise estimation layer & 4 & 4 & 4 \\
Min noise level $\beta_1$ & 0.0001 & 0.0001 & 0.0001 \\
Max noise level $\beta_T$ & 0.2 & 0.2 & 0.2 \\
Diffusion steps \textit{T} & 100 & 50 & 50 \\
Virtual nodes \textit{k} & 16 & 64 & 64 \\
\hline
\end{tabular}
\caption{The hyperparameters of CoFILL in each dataset}\label{tab:tab1parameter}
\end{table}

\subsection{Comparative Performance Study}\label{sec:experiment5result}

\begin{table*}[h!]
\scriptsize
\centering
\begin{tabular}{c|cc|cccc|cccc}
\hline
\multirow{3}{*}{} & \multicolumn{2}{c|}{\textbf{AQI-36}}                     & \multicolumn{4}{c|}{\textbf{METR-LA}}                                                                                       & \multicolumn{4}{c}{\textbf{PEMS-BAY}}                                                                                  \\ \cline{2-11} 
                  & \multicolumn{2}{c|}{SF}  & \multicolumn{2}{c|}{Block}                         & \multicolumn{2}{c|}{Point}    & \multicolumn{2}{c|}{Block}                       & \multicolumn{2}{c}{Point}  \\ \cline{2-11} 
                  & \multicolumn{1}{c|}{MAE}        & MSE           & \multicolumn{1}{c|}{MAE}        & \multicolumn{1}{c|}{MSE}         & \multicolumn{1}{c|}{MAE}        & MSE         & \multicolumn{1}{c|}{MAE}       & \multicolumn{1}{c|}{MSE}        & \multicolumn{1}{c|}{MAE}       & MSE        \\ \hline
Mean              & \multicolumn{1}{c|}{53.48±0.00} & 4578.08±0.00  & \multicolumn{1}{c|}{7.48±0.00}  & \multicolumn{1}{c|}{139.54±0.00} & \multicolumn{1}{c|}{7.56±0.00}  & 142.22±0.00 & \multicolumn{1}{c|}{5.46±0.00} & \multicolumn{1}{c|}{87.56±0.00} & \multicolumn{1}{c|}{5.42±0.00} & 86.59±0.00 \\
DA                & \multicolumn{1}{c|}{50.51±0.00} & 4416.10±0.00  & \multicolumn{1}{c|}{14.53±0.00} & \multicolumn{1}{c|}{445.08±0.00} & \multicolumn{1}{c|}{14.57±0.00} & 448.66±0.00 & \multicolumn{1}{c|}{3.30±0.00} & \multicolumn{1}{c|}{43.76±0.00} & \multicolumn{1}{c|}{3.35±0.00} & 44.50±0.00 \\
KNN               & \multicolumn{1}{c|}{30.21±0.00} & 2892.31±0.00  & \multicolumn{1}{c|}{7.79±0.00}  & \multicolumn{1}{c|}{124.61±0.00} & \multicolumn{1}{c|}{7.88±0.00}  & 129.29±0.00 & \multicolumn{1}{c|}{4.30±0.00} & \multicolumn{1}{c|}{49.90±0.00} & \multicolumn{1}{c|}{4.30±0.00} & 49.80±0.00 \\
Linear           & \multicolumn{1}{c|}{14.46±0.00} & 673.92±0.00   & \multicolumn{1}{c|}{3.26±0.00}  & \multicolumn{1}{c|}{33.76±0.00}  & \multicolumn{1}{c|}{2.43±0.00}  & 14.75±0.00  & \multicolumn{1}{c|}{1.54±0.00} & \multicolumn{1}{c|}{14.14±0.00} & \multicolumn{1}{c|}{0.76±0.00} & 1.74±0.00  \\
KF                & \multicolumn{1}{c|}{54.09±0.00} & 4942.26±0.00  & \multicolumn{1}{c|}{16.75±0.00} & \multicolumn{1}{c|}{534.69±0.00} & \multicolumn{1}{c|}{16.66±0.00} & 529.96±0.00 & \multicolumn{1}{c|}{5.64±0.00} & \multicolumn{1}{c|}{93.19±0.00} & \multicolumn{1}{c|}{5.68±0.00} & 93.32±0.00 \\
MICE              & \multicolumn{1}{c|}{30.37±0.00} & 2594.06±7.17  & \multicolumn{1}{c|}{4.22±0.05}  & \multicolumn{1}{c|}{51.07±1.25}  & \multicolumn{1}{c|}{4.42±0.07}  & 55.07±1.46  & \multicolumn{1}{c|}{2.94±0.02} & \multicolumn{1}{c|}{28.28±0.37} & \multicolumn{1}{c|}{3.09±0.02} & 31.43±0.41 \\
VAR               & \multicolumn{1}{c|}{15.64±0.09} & 833.46±13.85  & \multicolumn{1}{c|}{3.11±0.08}  & \multicolumn{1}{c|}{28.00±0.76}  & \multicolumn{1}{c|}{2.69±0.00}  & 21.10±0.02  & \multicolumn{1}{c|}{2.09±0.10} & \multicolumn{1}{c|}{16.06±0.73} & \multicolumn{1}{c|}{1.30±0.00} & 6.52±0.01  \\
TRMF              & \multicolumn{1}{c|}{15.46±0.08} & 1379.05±34.83 & \multicolumn{1}{c|}{2.96±0.00}  & \multicolumn{1}{c|}{22.65±0.13}  & \multicolumn{1}{c|}{2.86±0.00}  & 20.39±0.02  & \multicolumn{1}{c|}{1.95±0.01} & \multicolumn{1}{c|}{11.21±0.06} & \multicolumn{1}{c|}{1.85±0.00} & 10.03±00   \\
BATF              & \multicolumn{1}{c|}{15.21±0.27} & 662.87±29.55  & \multicolumn{1}{c|}{3.56±0.01}  & \multicolumn{1}{c|}{35.39±0.03}  & \multicolumn{1}{c|}{3.58±0.01}  & 36.05±0.02  & \multicolumn{1}{c|}{2.05±0.00} & \multicolumn{1}{c|}{14.48±0.01} & \multicolumn{1}{c|}{2.05±0.00} & 14.90±0.06 \\
V-RIN             & \multicolumn{1}{c|}{10.00±0.10} & 838.05±24.74  & \multicolumn{1}{c|}{6.84±0.17}  & \multicolumn{1}{c|}{150.08±6.13} & \multicolumn{1}{c|}{3.96±0.08}  & 49.98±1.30  & \multicolumn{1}{c|}{2.49±0.04} & \multicolumn{1}{c|}{36.12±0.66} & \multicolumn{1}{c|}{1.21±0.03} & 6.08±0.29  \\
GP-VAE            & \multicolumn{1}{c|}{25.71±0.30} & 2589.53±59.14 & \multicolumn{1}{c|}{6.55±0.09}  & \multicolumn{1}{c|}{122.33±2.05} & \multicolumn{1}{c|}{6.57±0.10}  & 127.26±3.97 & \multicolumn{1}{c|}{2.86±0.15} & \multicolumn{1}{c|}{26.80±2.10} & \multicolumn{1}{c|}{3.41±0.23} & 38.95±4.16 \\
rGAIN             & \multicolumn{1}{c|}{15.37±0.26} & 641.92±33.89  & \multicolumn{1}{c|}{2.90±0.0}   & \multicolumn{1}{c|}{21.67±0.15}  & \multicolumn{1}{c|}{2.83±0.01}  & 20.03±0.09  & \multicolumn{1}{c|}{2.18±0.01} & \multicolumn{1}{c|}{13.96±0.20} & \multicolumn{1}{c|}{1.88±0.02} & 10.37±0.20 \\
BRITS             & \multicolumn{1}{c|}{14.50±0.35} & 622.36±65.16  & \multicolumn{1}{c|}{2.34±0.01}  & \multicolumn{1}{c|}{17.00±0.14}  & \multicolumn{1}{c|}{2.34±0.00}  & 16.46±0.05  & \multicolumn{1}{c|}{1.70±0.01} & \multicolumn{1}{c|}{10.50±0.07} & \multicolumn{1}{c|}{1.47±0.00} & 7.94±0.03  \\
GRIN              & \multicolumn{1}{c|}{12.08±0.47} & 523.14±57.17  & \multicolumn{1}{c|}{2.03±0.00}  & \multicolumn{1}{c|}{13.26±0.05}  & \multicolumn{1}{c|}{1.91±0.00}  & 10.41±0.03  & \multicolumn{1}{c|}{1.14±0.01} & \multicolumn{1}{c|}{6.60±0.10}  & \multicolumn{1}{c|}{0.67±0.00} & 1.55±0.01  \\
CSDI              & \multicolumn{1}{c|}{9.51±0.10}  & 352.46±7.50   & \multicolumn{1}{c|}{1.98±0.00}  & \multicolumn{1}{c|}{12.62±0.60}  & \multicolumn{1}{c|}{1.79±0.00}  & 8.96±0.08   & \multicolumn{1}{c|}{0.86±0.00} & \multicolumn{1}{c|}{4.39±0.02}  & \multicolumn{1}{c|}{0.57±0.00} & 1.12±0.03  \\
PriSTI            & \multicolumn{1}{c|}{9.03±0.07}  & 310.39±7.03   & \multicolumn{1}{c|}{1.86±0.00}  & \multicolumn{1}{c|}{10.70±0.02}  & \multicolumn{1}{c|}{1.72±0.00}  & 8.24±0.05   & \multicolumn{1}{c|}{0.78±0.00} & \multicolumn{1}{c|}{3.31±0.01}  & \multicolumn{1}{c|}{\textbf{0.55±0.00}} & \textbf{1.03±0.00}  \\ \hline
CoFILL            & \multicolumn{1}{c|}{\textbf{8.70±0.06}}  & \textbf{296.52±7.29}   & \multicolumn{1}{c|}{\textbf{1.67±0.00}}  & \multicolumn{1}{c|}{\textbf{9.42±0.00}}   & \multicolumn{1}{c|}{\textbf{1.62±0.00}}  & \textbf{8.01±0.05}   & \multicolumn{1}{c|}{\textbf{0.74±0.00}} & \multicolumn{1}{c|}{\textbf{2.99±0.02}}  & \multicolumn{1}{c|}{0.57±0.00} & \textbf{1.03±0.00}  \\
Improvement           & \multicolumn{1}{c|}{3.65\%}     & 4.47\%        & \multicolumn{1}{c|}{10.22\%}    & \multicolumn{1}{c|}{11.96\%}     & \multicolumn{1}{c|}{5.81\%}     & 2.79\%      & \multicolumn{1}{c|}{5.13\%}    & \multicolumn{1}{c|}{9.67\%}     & \multicolumn{1}{c|}{---}       & ---        \\ \hline
\end{tabular}
\caption{The results of MAE and MSE for spatiotemporal imputation}\label{tab:tab2MAEMSE}
\end{table*}

\begin{table}[h!]
\footnotesize
\setlength{\tabcolsep}{3pt} 
\centering
\begin{tabular}{c|c|cc|cc}
\hline
\multirow{2}{*}{} & \textbf{AQI-36}        & \multicolumn{2}{c|}{\textbf{METR-LA}}                     & \multicolumn{2}{c}{\textbf{PEMS-BAY}}                     \\ \cline{2-6} 
                  & SF            & \multicolumn{1}{c|}{Block}        & Point        & \multicolumn{1}{c|}{Block}        & Point        \\ \hline
V-RIN             & 0.3154        & \multicolumn{1}{c|}{0.1283}       & 0.0781       & \multicolumn{1}{c|}{0.0394}       & 0.0191       \\
GP-VAE            & 0.3377        & \multicolumn{1}{c|}{0.1118}       & 0.0977       & \multicolumn{1}{c|}{0.0436}       & 0.0568       \\
CSDI              & 0.1056        & \multicolumn{1}{c|}{0.0260}       & 0.0235       & \multicolumn{1}{c|}{0.0127}       & 0.0067       \\
PriSTI            & 0.0997        & \multicolumn{1}{c|}{0.0244}       & 0.0227       & \multicolumn{1}{c|}{0.0093}       & {\textbf{0.0064}} \\ \hline
CoFILL            & {\textbf{0.09598}} & \multicolumn{1}{c|}{{\textbf{0.0220}}} & {\textbf{0.0213}} & \multicolumn{1}{c|}{{\textbf{0.0091}}} & 0.0069       \\ 
Improvement           & 3.73\%        & \multicolumn{1}{c|}{9.80\%}       & 6.17\%       & \multicolumn{1}{c|}{2.15\%}       & ---          \\ \hline
\end{tabular}
\caption{The results of CRPS for spatiotemporal imputation}\label{tab:tab3CRPS}
\end{table}

Our experimental evaluation demonstrates CoFILL's consistent superiority across multiple datasets and missing data scenarios. The quantitative results in Table~\ref{tab:tab2MAEMSE} reveal CoFILL's strong performance in terms of MAE and MSE metrics. For probabilistic imputation assessment, Table~\ref{tab:tab3CRPS} presents the CRPS comparisons. CoFILL achieves the best performance in 12 out of 15 experimental configurations spanning different metrics and missing patterns. The improvement becomes particularly apparent when compared with PriSTI, the state-of-the-art method. On the METR-LA dataset with Block, CoFILL reduces both MAE and MSE by 10.22\%. On the PEMS-BAY point-missing pattern, CoFILL performs on par with PriSTI (only $\leq $1\% difference with identical MSE).

The comparative analysis reveals clear performance patterns across different methodological categories. Traditional statistical and machine learning approaches, such as Mean and KNN, show limited effectiveness due to their simplified assumptions about spatiotemporal relationships. The deep learning approaches, represented by GRIN and BRITS, face performance degradation from accumulated errors in their autoregressive processing. While recent diffusion-based models like CSDI and PriSTI demonstrate strong baseline performance, CoFILL still outperforms them with advanced conditional information construction and feature fusion methods.


\subsection{Ablation Study}\label{sec:experiment6ablation}

We evaluate the impact of key components of CoFILL by comparing the following settings.

\begin{itemize}
    \item w/o Forward (Forward Interpolation): Pre-imputation information \( X_1 \) is removed, and conditional feature extraction is performed directly on the missing data.
    \item w/o Temporal (Temporal Domain): The temporal module in the conditional feature extraction block is removed, leaving only the frequency domain module.
    \item w/o Frequency (Frequency Domain): The frequency module in the conditional feature extraction block is removed, leaving only the temporal module.
    \item w/o Cross (Cross Attention): The cross-attention mechanism between temporal and frequency domains is removed and replaced by direct addition.
\end{itemize}

Table~\ref{tab:tab5ablation} presents the impact of each setting on the performance. 
We only analyze AQI-36 and METR-LA due to the computational cost of re-running diffusion-based CoFILL, especially on the larger PEMS-BAY.
This aligns with the experimental setup of the baseline PriSTI~\cite{ref22}.
The removal of forward interpolation produces the most significant performance degradation across datasets. For AQI-36, this modification increases the MAE from 8.7 to 9.15, representing a substantial 0.45 unit deterioration. The temporal domain component emerges as the second most critical element, with its removal causing more severe performance decline than eliminating the frequency domain component from the conditional information module described in Section~\ref{sec:method2}. The cross-attention mechanism shows the least impact on model performance, particularly evident in the METR-LA dataset where its removal results in minimal MAE changes.

\begin{table}[h!]
\footnotesize
\centering
\begin{tabular}{c|c|cc}
\hline
\multirow{2}{*}{} & \textbf{AQI-36}            & \multicolumn{2}{c}{\textbf{METR-LA}}                        \\ \cline{2-4} 
                  & SF & \multicolumn{1}{c|}{Block} & Point \\ \hline
w/o Forward            & 9.15              & \multicolumn{1}{c|}{1.95}          & 1.83          \\
w/o Temporal            & 9.00              & \multicolumn{1}{c|}{1.89}          & 1.78          \\
w/o Frequency            & 8.81              & \multicolumn{1}{c|}{1.71}          & 1.72          \\
w/o Cross            & 8.82              & \multicolumn{1}{c|}{1.68}          & 1.63          \\ \hline
CoFILL            & \textbf{8.70}              & \multicolumn{1}{c|}{\textbf{1.67}}          & \textbf{1.62}          \\ \hline
\end{tabular}
\caption{The result of MAE for ablation studies}\label{tab:tab5ablation}
\end{table}


\subsection{Hyperparameter Sensitivity Analysis}\label{sec:experiment7hyper}

This experiment analyzes two important parameters in the diffusion model: max noise level ($\beta_T$) and channel size of the hidden state ($d$) that represent fundamental trade-offs. The former controls the intensity of the noise added during the diffusion process. A high noise level ($\beta_T$) may lead to distorted generated data, while a low noise level ($\beta_T$) may make it difficult to learn the various relationships within the data. The latter determines the model's ability to represent and handle information at each time step. 
A larger $d$ allows the model to learn more features, but it may also lead to higher overhead. 

\begin{figure}[t]
	\centerline{\includegraphics[width=1\linewidth]{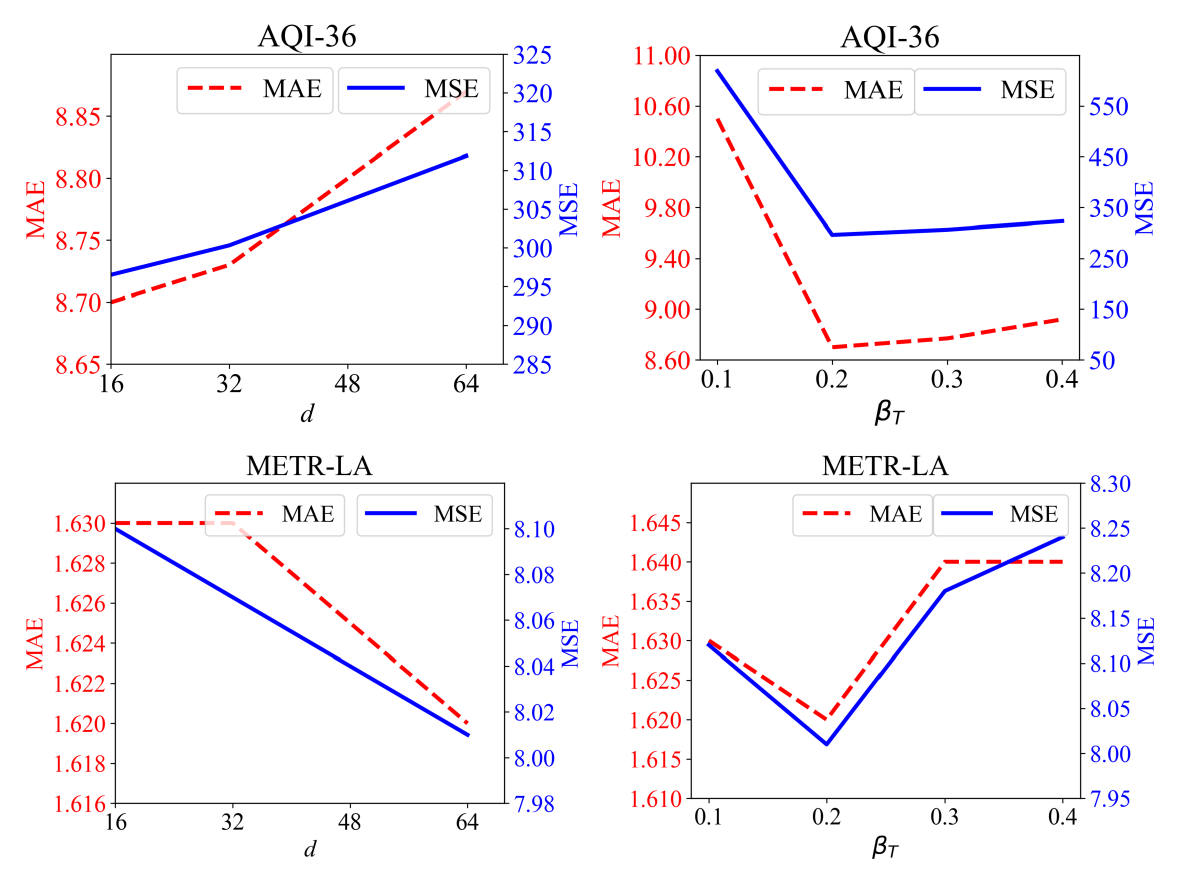}}
	\caption{Analysis of hyperparameters $\beta_T$ and $d$.}
	\label{fig:fig3experiment}
\end{figure}

The experimental results are illustrated in Figure~\ref{fig:fig3experiment} (only for AQI-36 and METR-LA due to the same reasons discussed in Section~\ref{sec:experiment6ablation}). For AQI-36, the model maintains stable MAE and MSE values with maximum noise level $\beta_T$ between 0.2 and 0.4. The METR-LA dataset shows higher sensitivity to $\beta_T$ variations within this range. The channel size $d$ produces opposing effects, including degrading performance on AQI-36 (increasing $d$) but improving results on METR-LA. This behavior reflects the different complexities in air quality and traffic data.
Furthermore, we set $d$ to 16 for AQI-36 and 64 for METR-LA. We apply METR-LA's parameter to PEMS-BAY due to their similar data characteristics, as detailed in Table~\ref{tab:tab1parameter}. This enables CoFILL to outperform PriSTI across most scenarios, which obtains better performance even under PEMS-BAY's Point where the metrics remain close to PriSTI.

\section{Conclusion}

In this paper, we present CoFILL, a novel conditional diffusion framework for spatiotemporal data imputation. CoFILL first overcomes the error accumulation problem by implementing a non-recursive processing structure through our diffusion framework. Second, it resolves the challenge of capturing complex temporal dependencies by introducing a dual-stream architecture that simultaneously processes temporal and frequency domain features. Extensive experiments on real-world datasets demonstrate CoFILL's superior performance. Our model achieves better imputation accuracy than SOTAs across various missing data scenarios.
We will develop adaptive fusion and extend CoFILL to handle multi-resolution spatiotemporal data in the future.

\section*{Acknowledgments}

This work is supported by the National Natural Science Foundation of China (62302148), Hebei Natural Science Foundation (F2024202076), and Hebei Yanzhao HuangJintai Talents Program (Postdoctoral Platform) (B2024003002).


\bibliographystyle{named}

\end{document}